\documentclass[10pt,twocolumn]{article}
\usepackage{cite}

\usepackage[pdftex]{graphicx}
\usepackage{amsmath,amssymb,amsthm}
\usepackage{algorithmic}
\usepackage[caption=false,font=footnotesize]{subfig}
\usepackage{url}

\usepackage{color}
\usepackage{authblk}

\usepackage{glossaries}
\newacronym{ml}{ML}{Machine Learning}
\newacronym{fl}{FL}{Federated Learning}
\newacronym{an}{AD}{Anomaly Detection}
\newacronym{rf}{RF}{Random Forest}
\newacronym{bc}{BC}{blockchain}

\newcommand{\R}{\mathbb{R}}
\newcommand{\N}{\mathbb{N}}
\DeclareMathOperator{\Ens}{Ens}
\DeclareMathOperator{\Estim}{Estim}
\newcommand{\share}{\mathrm{share}}
\newcommand{\new}{\mathrm{new}}
\newcommand{\node}{\mathrm{node}}
\newcommand{\tree}{\mathrm{tree}}
\newcommand{\feat}{\mathrm{feat}}

\title{A Framework for Verifiable and Auditable Federated Anomaly Detection}

\author[1]{Gabriele Santin\thanks{gsantin@fbk.eu}}
\author[2]{Inna Skarbovsky\thanks{inna@il.ibm.com}}
\author[2]{Fabiana Fournier\thanks{fabiana@il.ibm.com}}
\author[1]{Bruno Lepri\thanks{lepri@fbk.eu}}
        
\affil[1]{Digital Society Center, Bruno Kessler Foundation (FBK), Trento, Italy}
\affil[2]{IBM Research, Haifa, Israel}

\begin{document}

\maketitle

\begin{abstract}
Federated Leaning is an emerging approach to manage cooperation between a group of agents for the solution of Machine Learning tasks, 
with the goal of improving each agent's performance without disclosing any data.

In this paper we present a novel algorithmic architecture that tackle this problem in the particular case of Anomaly Detection (or classification or rare 
events), a setting where typical applications often comprise data with sensible information, but where the scarcity of anomalous examples encourages
collaboration.

We show how Random Forests can be used as a tool for the development of accurate classifiers with an effective insight-sharing  mechanism that does not break 
the data integrity.
Moreover, we explain how the new architecture can be readily integrated in a blockchain infrastructure to ensure the verifiable and auditable execution of the 
algorithm.

Furthermore, we discuss how this work may set the basis for a more general approach for the design of federated ensemble-learning methods beyond the specific 
task and 
architecture discussed in this paper.

\end{abstract}

\section{Introduction}\label{sec:introduction}

In a data-driven world, \gls{ml} has progressively established itself as a fundamental tool that spreads across multiple fields and permeates an 
increasing variety of applications. 
After a decade of fast technological developments mainly driven by the exceptional new results achieved by Deep Learning \cite{LeCun2015,Goodfellow2016}, a new wave of reflection is emerging about the scope, applicability, and technical limitations of these techniques.
In particular, an increasing new attention is devoted to the issues of data ownership, data privacy, and data trading. 

In this setting, multiple related aspects are being analyzed and systematized within the framework of \gls{fl} 
\cite{McMahan2017,Zhang2021,Horvath2021,Yang2019}, and several real-world problems have been approached with these techniques, e.g. in the banking \cite{Long2020,Yang2019b} and health \cite{Dayan2021, Sheller2020} sectors, even beyond classical domains \cite{Liu2022}, and considering privacy and fairness constraints \cite{Shi2021,Kang2021}.
This new field deals with the study of various scenarios where multiple agents own separate batches of data, and they are willing to 
cooperate for the construction of some \gls{ml} models. This collaboration leverages different communication strategies to overcome the 
limitations of the single agents, which can be due to scarcity of data or scarcity of computational resources, but with the important constraint that data 
should never leave the location where it resides. This approach is in stark contrast with more traditional data-centralized methods, and it paves the way for a 
number of new algorithms that focus on various aspects of data ownership.

For a comprehensive analysis of the key goals, applications, and challenges of \gls{fl} we refer to the recent overviews \cite{Yang2019, Kairouz2021,Yang2019c}. To put our approach 
into context, we just recall that there is an important distinction between centralized and decentralized \gls{fl} \cite{Abdulrahman2020, Kairouz2021,Diao2020}. 
In the first case, a central orchestrator coordinates a set of distributed agents (or nodes) and their computational resources to improve the fitting of a 
central model. 
In the second case, instead, the entire process is collectively directed by the distributed agents. Heterogeneous cases are also of interest, where the central 
controller acts, or is queried, only when needed. In all cases, the focus of current research are the issues related to communication 
efficiency, to the influence of the topology of the connections in the agents' network, and the quality of the learned model. Additionally, in the decentralized 
approach the absence of an omniscient orchestrator opens the way for new possibilities for privacy preservation and flexibility, but it poses new challenges for 
the security of the communications, the integrity of the system, and the accuracy of the algorithmic procedure. 

In this paper we focus on \gls{an} \cite{Chandola2009,Ahmed2016} as a use case for \gls{fl}. 
The scenario is motivated by \gls{an} systems that are common in the financial industry, such as fraud detectors or default predictors. 
The peculiar characteristic of these applications is that a classifier has to be trained to identify anomalous cases, i.e., events that are unusual compared to 
the most frequent patterns observed in the data. In particular, anomalous examples are scarce by definition. 
As a consequence, different agents such as banks, financial institutions, insurance companies may foresee a benefit in collaborating with their peers in order 
to trade knowledge and improve their individual models. 
On the other hand, the data that is used to train these systems is usually shared with caution, since it typically comprises sensitive personal 
information regarding the financial position or the individual characteristics of the clients. Moreover, the possession of these data is often an 
important asset for the single agents, which are possibly not willing to give them away once for all, but would rather like to develop an on-purpose sharing. 
This option is inherently difficult with easily copyable digital data. 

With these constraints and goals in mind, we present in this paper a fully decentralized \gls{fl} system where multiple agents collaborate for the training of 
one model per agent, and which is privacy preserving by design, robust to changes in the network topology and to asynchronous communications, and resistant to 
malicious intrusions and adversarial attacks. 

The system is designed so that each agent trains an ensemble classifier \cite{Sagi2018}, i.e., a \gls{ml} model that is made of multiple simple estimators that are combined as 
atomic building blocks. This structure makes it easy to iteratively improve local models as well as exchanging knowledge between agents by sharing the top 
performing blocks. We use in particular \glspl{rf} \cite{Breiman2001} as ensemble models, as they are well-suited for anomaly detection problems and robust to missing data, but 
we comment along the paper how this is not a restrictive choice and other ensembles could be adopted. 
Moreover, the chosen design of the \gls{ml} algorithm permits to integrate the system in a \gls{bc} infrastructure that guarantees trustable and verifiable 
execution of the algorithm, and certifies the communication between the nodes. 

Other works have proposed solutions for the integration of \gls{fl} in a \gls{bc} environment \cite{Pokhrel2020,Ma2020,Li2021,Wang2021,Xu2021,Rehman2020,Li2021x}. In this work, we introduce two main novelties over existing approaches: 
(i) The framework supports federations where the agents are connected by means of a time-varying network in a fully decentralized scenario. This includes the case of single agents joining or leaving the group at different times, or exploiting the collaboration in an on-demand fashion. 
This opens the way for treating the participation in the federation as a tradable utility (see Section \ref{sec:federated} and Section \ref{sec:verify}), and leverages the \gls{bc} as a verification tool; 
(ii) The solution is algorithm-agnostic in its main components, meaning that it can be applied on top of a large class of machine learning models, provided that some atomic operations can be defined (see Section \ref{sec:atomic}). 
In particular, the algorithm is not bound to specific architectures or optimization methods.

The paper is organized as follows. We start by recalling the necessary background on \gls{rf} and \gls{bc} in Section~\ref{sec:background}, and with these 
tools we introduce the novel algorithm in Section~\ref{sec:algorithm} and discuss the full \gls{bc} solution in Section~\ref{sec:blockchain}. 
We validate our new system through a number of experiments in Section~\ref{sec:experiments}, and conclude by discussing some perspectives and open problems in 
Section~\ref{sec:conclusions}.

\section{Background}\label{sec:background}
We start by recalling some background details in order to facilitate the reading of the paper by researchers from both the 
\gls{ml} and \gls{bc} communities.

\subsection{Setting of the \gls{ml} algorithm}\label{sec:setup_ml}
In the following we assume that each agent has a labeled dataset of examples, where each data point (e.g., a transaction) is represented by a 
$d$-dimensional vector $x:=(x^{1},\dots, x^{d})\in\R^d$, collecting $d$ features $x_i$ (e.g., the ID of the user performing the transaction, its timestamp, the 
amount transferred, etc.). Each example is associated to a label $y_i\in\{0,1\}$ indicating whether the $i$-th example is normal ($y_i=0$) or anomalous 
($y_i=1$).
These examples are collected in a dataset $(\mathcal X, \mathcal Y)$ of $m\in\N$ data points $\mathcal X:=\{x_1, \dots, x_m\}$ with labels $\mathcal 
Y:=\{y_1,\dots, y_m\}$.
In this paper we work with tabular data, but this is not required in general and other data types may be supported, such as images or texts.

For the detection of anomalous examples each agent trains its own classifier, i.e., a map $\Phi:\mathcal X\to [0,1]$ that is 
optimized on the training set, and that can be used to approximately predict the class of an unseen data point $x$, with the usual convention that the example 
is classified as normal if $\Phi(x)\leq 0.5$ and as anomalous if $\Phi(x) > 0.5$. 
We consider ensemble classifiers, which means that we actually train a set of $n\in\N$ simpler classifiers (or estimators) $\phi^i:\mathcal X\to [0,1]$, $1\leq 
i\leq n$, each trained on the same classification task, and define the global prediction of $\Phi$ either by averaging, i.e., 
$\Phi(x):=\frac1n\sum_{i=1}^n\phi^i(x)$, or by majority voting among the $n$ predictions $\{\phi^i(x)\}_{i=1}^n$. 
To explicitly denote the transformation from the ensemble to the estimators and vice-versa, we use the notation $\Phi:=\Ens\left(\{\phi^i\}_{i=1}^n\right)$ 
and $\{\phi^i\}_{i=1}^{n}:=\Estim(\Phi)$.
This kind of classifiers will be instrumental for our construction, since they are quite straightforward to improve by enlarging the ensemble size $n$ and 
adding new simple learners, and it is possible to mix different classifiers $\Phi$ and $\Phi'$ by mixing their simple learners.

As a prototype of ensemble classifiers, in this paper we focus on \glspl{rf} \cite{Ho1995}, which use decision trees as their simple learners. 
Decision trees \cite{Breiman2017} are maps $\phi^i:\mathcal X\to[0,1]$ that compute their prediction according to a binary tree: Once the tree is 
trained on the data, at prediction time an input enters the tree from its root, and it follows a sequence of binary tests until it reaches a leaf 
node. Each of these leaf nodes is associated to a unique label, which is the prediction assigned by the tree to each input that falls into this leaf. At each 
non-leaf node, instead, the splitting is decided by the value of a single feature of the input, and thus a decision tree can be understood as a sequence of 
binary splits of the input space according to a subset of features at given splitting values. The training of this structure requires to select the sequence of 
features and the threshold values to define the splitting, and this is usually realized by guaranteeing that the examples in the training set are 
distributed in a balanced manner among the leaf nodes, and adopting criteria for the growth of the tree in depth and width. We refer to \cite{Breiman2001} for 
a detailed treatment of this topic. 

In addition to their basic ensemble structure, \glspl{rf} perform two randomization operations to improve their accuracy and robustness. 
Namely, \glspl{rf} are trained by bootstrap aggregation, i.e., each tree in the ensemble is trained on a random subset of the full dataset, extracted  
by a sampling with replacement. Additionally, the single trees are trained with feature bagging, i.e., each splitting of each tree is constructed by considering 
only a uniformly randomly selected subset of the features of the data. 

\glspl{rf} are particularly suited for tabular data and they can deal quite effectively with missing entries thanks to their structure that do not require the 
knowledge of each single feature. Moreover, their training is quite simple and thus suitable to be performed repeatedly, as will be the case in our algorithm.

\subsection{Setting of the \gls{bc} solution}
A \gls{bc} is essentially a digital ledger of transactions that is duplicated and distributed across the participants in the \gls{bc} network. 
Transactions are recorded in a final and immutable manner by the \gls{bc}, providing all network members with an identical and trustworthy real-time view of 
the state. Due to its inherent characteristics, \gls{bc} is the natural platform to support privacy and trust as well as a secure execution environment \cite{Guo2022,Meng2018,Christidis2016}. 
Our proposed BC solution ensures a secure, auditable, and verifiable framework for execution of federated learning algorithms. 

The idea is that each learning node in the BC network publishes intermediate results 
at the end of each iteration. 
These results can be consumed by other learning nodes to improve the accuracy of their next computations. 
Our solution is generic and can support any \gls{ml} algorithm having the following properties: The algorithm can be represented as a portable computation 
workload (e.g., a docker image which can be instantiated to a container running the algorithm’s computation); the algorithm can be iterative or single-step; and 
it can either be centralized and require orchestration and synchronization between iterations or be distributed and thus self-orchestrating. 

For our proposed framework, as underlying \gls{bc} technology we leverage Hyperledger Fabric (or simply Fabric) \cite{Androulaki2018,FabricWeb}, which is one 
of the most promising \gls{bc} platforms for enterprises (see e.g., \cite{Gupta2018} for a comprehensive 
and foundational analysis of the \gls{bc} solutions and services for enterprises). 



\section{Federated training of ensemble classifiers}\label{sec:algorithm}
With these tools in hand we now introduce the federated learning algorithm. We will first formulate the algorithm under as general assumptions 
as possible, and then we provide some specifications in the case of \glspl{rf}. We will anyhow comment on how these can be generalized to different scenarios.

\subsection{Agents and atomic operations}\label{sec:atomic}
We assume to have a number $N\in\N$ of agents (or nodes) participating in the federation, and denote them as $V:=\{v_1, \dots, v_N\}$. Each node $v_j$ has an 
own dataset $(\mathcal X_j, \mathcal Y_j)$ of size $m_j$ of the form described in Section \ref{sec:setup_ml}, and its goal is to obtain an 
ensemble classifier $\Phi_{j}$ for the detection of anomalies, working possibly beyond its own data. 

We consider three atomic operations to modify an ensemble: one enlarges the ensemble, one keeps its size bounded, 
and one selects the top performing estimators. Assuming that $\Phi$ is an existing ensemble with $n$ estimators, $\{\phi^i\}_{i=1}^{n'}$ is another 
set of estimators, and $k\in\N$ is an integer parameter, the three operations are formally defined as follows:
\begin{itemize}
\item \texttt{ADD}$(\Phi, \{\phi^i\}_{i=1}^{n'})$ returns the enlarged ensemble $\Phi':=\Ens\left(\Estim(\Phi)\cup 
\{\phi^i\}_{i=1}^{n'}\right)$.
\item \texttt{GET\_TOP}$(\Phi, k)$ sorts the $n$ estimators of $\Phi$ according to some order that needs to be specified, and returns the 
top $k$. If $n\leq k$ all the $n$ estimators are returned.
\item \texttt{CROP}$(\Phi, k)$ keeps only the $k$ best estimators of an ensemble $\Phi$, i.e., it sets $\Phi=\Ens($\texttt{GET\_TOP}$(\Phi, 
k))$.
\end{itemize}

\subsection{Federated learning}\label{sec:federated}
The group of agents is partially connected according to a network represented by an undirected graph $G=(V, E)$, where there is an edge 
$(v_i, v_j)\in E$ if and only if a connection is active between the $i$-th and $j$-th node. 

The assumption that the connection graph $G$ is fixed is only made for simplicity of exposition, but it is straightforward to deal with 
time-varying graphs that may represent e.g. agents entering and leaving the federation, or temporary failures in the connection system. Indeed, for the 
algorithm to run it is sufficient to assume that each node $v_j$, whenever it is interested in a communication, is able to get the list of its first order 
neighbors, i.e., the set of all agents $v_i$ 
such that there is a link $(v_j, v_i)\in E$. 
Moreover, each node in practice has no need to know the entire graph, and has no option to modify it. More advanced 
scenarios could be envisioned and investigated, for example by assigning to the agents a certain budget that can be used to establish optimized connections to certain nodes, or by using the knowledge of the entire connection graph to take some decision on the learning mechanism. We leave these extensions for future work.

To manage the communication, each node $v_j$ has a registry $R_j$ with a slot $R_{j}(v_i)$ for each of the other nodes $v_i$. We assume that each 
node $v_i$ can write a message to the slot $R_{j}(v_i)$ in the registry of the node $v_j$ if this is one of its first order neighbors. 

Using the registry and the atomic operations on the ensemble, we are in the position to define the three fundamental operations that each agent 
$v_j$ can perform to change its status at each iteration. They are controlled by three parameters $n_{\new}, n_{\max}, n_{\share}\in\N$ that 
we assume to be globally set, even if local parameters (i.e., node-dependent) may be used without significant modifications. The three operations are the 
following:

\begin{enumerate}
\item \texttt{FIT}: A number $n_{\new}\in\N$ of simple learners $\{\phi^i\}_{i=1}^{n_{\new}}$ are trained by the agent on its own dataset $(\mathcal X_j, 
\mathcal 
Y_j)$, and the ensemble $\Phi_j$ is enlarged as $\Phi_j:=$\texttt{ADD}$(\Phi_j, \{\phi^i\}_{i=1}^{n_{\new}})$. If the resulting number of estimators is larger 
than $n_{\max}$, then the method \texttt{CROP}$(\Phi_j, n_{\max})$ is used to keep only the best ones. 
\item \texttt{SHARE}: The agent identifies its top $n_{\share}$ estimators, and writes them to the registry of each of its first order neighbors. If a registry 
slot contains already some estimators from previous communications, they are overwritten.
\item \texttt{GET}: The agent reads its registry slots to collect all the estimators received in the previous iterations (if any), and adds them to its current 
ensemble by 
using the \texttt{ADD} method. If this operation makes the ensemble larger than $n_{\max}$, excess estimators are removed by a call to the \texttt{CROP} method.
\end{enumerate}

Finally, the algorithm requires initialization and termination conditions. For simplicity we assume that each 
agent $v_j$ starts with an empty ensemble $\Phi_j:=\Ens(\emptyset)$ and runs \texttt{FIT} as its first operation. Moreover, each agent
terminates its execution when the prescribed iterations are executed.

\subsection{Properties of the algorithm}
The entire algorithm is completely decentralized, since it only requires the existence of a communication network and the agreement 
on a set of initial parameters. The model supports time-varying networks, and it allows for completely asynchronous communication, including the option for 
different nodes to join or leave the federation at different times.

Observe that all the operations except for \texttt{GET\_TOP} are well defined for any type of ensemble classifier, and do 
not require further specification to be implementable. The only method-specific operation is thus \texttt{GET\_TOP}, that requires to define a way to rank 
the estimators within an ensemble. We discuss our solution in the case of \glspl{rf} in the next section, but we remark that this choice is not unique, and 
that similar design principles could be adopted to work with more general ensembles. In this sense, the present algorithm may be understood as a family of 
algorithms, parametrized by the method that is used to promote some estimators with respect to other ones.

The importance of this ranking system is reflected in the fact that we are employing a registry with slots that stores only the last written information. In 
this way, when a node reads its registry via the \texttt{GET} method, it only reads the result of the most recent call of \texttt{GET\_TOP} transmitted by its 
neighbors. 

This solution is used also to guarantee that the registry has bounded memory footprint, since in this way it needs to store at most $n_{\share}\cdot N$ 
estimators at 
each time. Similarly, the bound $n_{\max}$ on the number of estimators held by each single node controls the size of each ensemble classifier. These two 
requirements can be translated to memory bounds if we assume that each estimator has a maximal memory size.

Moreover, the only operation that can create new estimators is \texttt{FIT}. Whenever this method is called, 
the newly constructed estimators are labeled with identifiers $(v_j, i)$, where $v_j$ is the identifier of the creator node, and $i$ is a progressive counter 
maintained by $v_j$. In this way each estimator in the federation is uniquely identified, and it is always possible to know which nodes trained it.
Moreover, communication between different nodes amounts only at the exchange of estimators via the \texttt{SHARE} and \texttt{GET} methods. Both the operations 
of creation and sharing are thus easily secured by means of the \gls{bc} integration that we are discussing in detail in Section~\ref{sec:blockchain}, so that 
the federation is protected against anomalous agents and malicious injections of information.

\subsection{Ranking of the estimators for \gls{rf}}
To obtain a fully functioning algorithm, it remains to specify the mechanism used to rank the estimators within each ensemble, i.e., to define the 
\texttt{GET\_TOP} operation. We define it for \glspl{rf}, which are the method of choice of this paper.

As discussed before, the sorting of the estimators is the most delicate operations and the one that have the largest potential to affect the result of 
the 
algorithm. In general terms, we aim at using unsupervised methods for this task, namely, we do not use the labels of the data to sort the estimators. The 
reason for this 
choice is that any supervised operation must rely on the data available to each node, and using the same local data that are used for training to rank the 
estimators is very likely to lead to a downplay of the importance of the estimators received from the other nodes.
For this reason, we decided to analyze only methods that rely on the structure of the estimators. 

Although different \gls{rf} pruning schemes have been introduced \cite{Kulkarni2012,Giffon2020, Feng2016}, we use here a mechanism that allows us 
to obtain a full sorting of the set of trees, and not only a reduction of its number. To this end, we recall that each estimator is a decision 
tree, and thus it can be represented by a tree where each non terminal node $v$ is associated with the index $s(v)\in\{1, \dots, d\}$ of the splitting 
feature, and the corresponding splitting value $x(v)\in\R$ (see section \ref{sec:setup_ml}). We use the splitting index $s(v)$ to identify the type of a node, 
and we regard $x(v)$ as node feature, so that each decision tree can be identified as $D:=(T, X)$, where $T$ is a tree with labeled nodes, and $X$ is a 
vector of node features associated to the non-terminal nodes.

Given a pair of decision trees $D:=(T, X)$, $D':=(T', X')$, we define a similarity measure that is used to compute the estimators' ranking in a 
structure-dependent way, i.e., one that takes into account the definition of each single estimator. 
To this end we define a positive definite and symmetric kernel $k(D, D')$ over pairs of decision trees. The kernel is  
a modification of the tree kernel of \cite{Haussler1999}, and we provide its explicit construction in Section~\ref{sec:kernel}.
We refer to \cite{Shawe-Taylor2004,Wendland2005} for a detailed treatment of the topic of kernel methods, and we recall here that $k$ can be used to encode 
general data (decision trees in this case) in a possibly high dimensional Hilbert space where standard numerical techniques are available. Moreover, the same method can be extended to other ensembles as soon as a kernel can be defined on its building blocks, and thus the present method has the potential to be applied in more general settings.

In particular, it is possible to define a Gaussian Process \cite{Rasmussen2006} with covariance function $k$ over the space of decision trees. Given the 
process, one may select a subset of the set of trees so that, conditioning the process on the labels associated to these trees, 
the maximal standard deviation of the posterior process is minimized. In this sense, this subset of trees may be regarded as the one that control the maximal 
variation in the data. This problem may be efficiently approximated by a greedy algorithm \cite{DeMarchi2005} that selects this set in an iterative way, and 
this gives the ordering of the estimators that we are looking after. It can be shown that this process is quasi-optimal \cite{Santin2017,Wenzel2021a}, meaning 
that the greedy selection is as effective as a global optimization, up to a constant. Running this algorithm until it selects $k$ elements, we obtain 
an ordered sequence $D_1, \dots, D_{k}$ representing the $n$ most important estimator, thus implementing the \texttt{GET\_TOP} operation.

\section{A Blockchain solution for secure and trustworthy federated learning}\label{sec:blockchain}
We describe now in detail our proposed framework, and we refer to \cite{FabricWeb,Baset2018} for more details on 
Hyperledger Fabric.

\subsection{Structure of the blockchain solution}
\begin{figure}
\includegraphics[width=.5\textwidth]{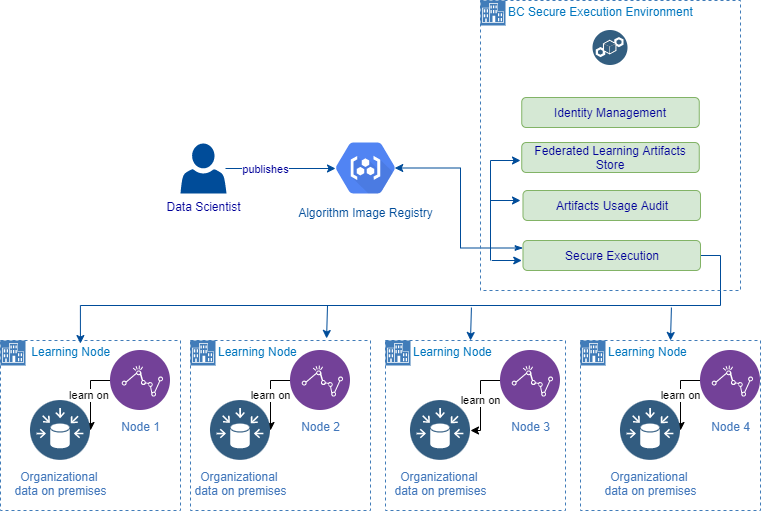}
\caption{Verifiable and auditable federated machine learning framework}
\label{fig:bc_figure_1}
\end{figure}

The framework consist of different conceptual elements (see Figure~\ref{fig:bc_figure_1}): 
(i) a Data Scientist, responsible for creating and pushing the federated learning algorithm image to the algorithm image registry after the training phase of 
the algorithm is over; (ii)
the Algorithm image registry,  which is any kind of local or hosted image registry for storage of docker images representing the ML 
algorithms; and (iii)
the Learning Nodes, which are the organizational nodes of organizations in the \gls{bc} network participating in the federated learning process. Here, production data is stored in 
premises and only intermediate and final results of the algorithm execution are stored in the \gls{bc} ledger.

Additionally, the system comprises a \gls{bc} execution environment, i.e., a secure execution environment that provides a verifiable privacy-preserving 
computation environment for federated learning scenarios. The environment comprises the following modules:

\begin{itemize}
\item Identity Management:  A built-in service in Hyperledger Fabric that provides a membership identity that manages user IDs and authenticates all participants in the 
network including (i) the specification of Certification Authority (CA) servers (defined as part of the \gls{bc} network configuration), (ii) the 
certification of users and applications using these CA servers, and (iii) mechanisms to sign and validate the signatures of all transactions and messages 
submitted to the network.

\item Federated Learning Artifacts Store: The chaincodes implementing the business logic for storing, updating, retrieving, and querying business artifacts 
related to federated learning, i.e., algorithm images' metadata, metadata of the learning process, and intermediate results and models.

\item Artifacts Usage Audit: The inherent functionalities in chaincodes which allow to query the history of updates for each artifact stored in the ledger, 
thus allowing to present a clear and complete picture of the artifact’s provenance.

\item Secure Execution: This module securely runs the computation tasks of the ML algorithm (we refer to computation task or workload as the ML algorithm 
instance or iteration), producing signed outputs (i.e., the insights from the learning round), and storing these outputs in the ledger. In the case of 
federated learning, it helps to establish the auditability and verifiability of the execution of local ML models and to improve the trust among the 
participants. Moreover, in the case of updates to the learning algorithm, it is guaranteed that all the parties are aware of the correct image version and are 
enforced to use the correct one to participate in the learning process.
\end{itemize}

\subsection{Verifiability of the execution}\label{sec:verify}

Our proposed approach allows delegating the computation over sensitive data to the data owner, while establishing trust of the rest of the stakeholders in the 
computation result. This is achieved via implementation of the following core characteristics:
\begin{itemize}
\item The computation workload is portable so that it is possible to deploy it in the data owner's environment.
\item The integrity of the computation workload is verifiable, i.e., computation stakeholders have guarantees that the actual computation was performed on the 
respective data.
\item The provenance over the input data, the output of the computation, and the computation logic is tracked.
\end{itemize}

We implement the portability characteristic by packaging the computation logic in a portable artifact. A docker image is an example of such a portable 
artifact, which is suitable for relatively simple computations that allow incorporating the entire logic into a single image. In cases where the computation 
involves multiple steps and components, it can be packaged as a composite asset, consisting of a set of images (each incorporating a relevant phase or function 
in the computation) and an artifact (or a set of artifacts) that define the orchestration and the choreography of the composite computation.

To establish correctness and integrity guarantees over the computation logic, we propose to manage computation workloads metadata in \gls{bc}. Having the 
metadata record in a shared distributed ledger ensures that all the parties have joint understanding of how to verify that a given portable deployable artifact 
is of the correct version and its contents have not been tampered with. For the algorithm images, we store a SHA256 hash of the docker image on the ledger. At 
the time of computation task creation from an image, when pulling the image from the algorithm image registry, we can verify image authenticity by calculating 
and comparing the image’s hash to the one stored in the ledger. For the computation task results we use public/private key verification. When creating a 
computation task, we use a crypto library to generate private/public key pairs. A public key of the pair is stored in the ledger in the execution task record, 
while the private key is passed to the computation task runtime. Once it finalizes, the computation task updates its record in the chain with the results of the 
execution signed with the private key. The updating chaincode then verifies the signed result element with the public key of the computation task to ensure that 
the results are being updated by the entity with the correct private key. 

Trackability and provenance is gained by providing auditing and verifiability capabilities for managing the \gls{ml} algorithm image lifecycle (e.g., publishing 
a 
new algorithm image and usage of the algorithm image), for secure execution (ensuring, for example, that the correct algorithm image is used in each execution 
of computation task), and for recording of intermediate results (allowing to answer questions, such as which artifacts were published at the end of each run 
for a particular learning process, or what artifacts a particular organization published for a particular learning task).

\begin{figure}
\includegraphics[width=.5\textwidth]{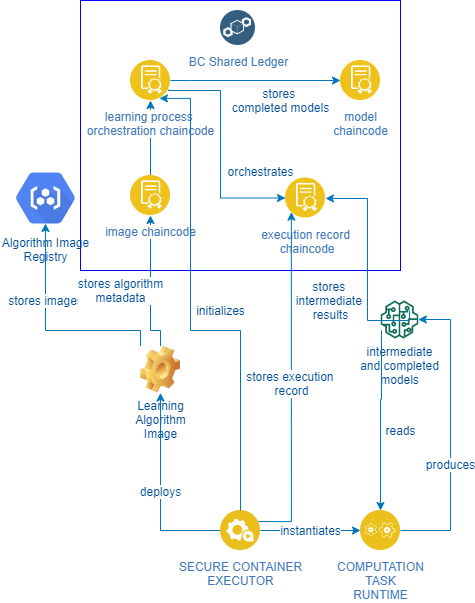}
\caption{Blockchain solution building blocks and flows.}
\label{fig:bc_figure_2}
\end{figure}

Figure~\ref{fig:bc_figure_2} depicts the interactions between the chaincodes in our BC network and the other building blocks comprising our secure execution 
environment. These 
building blocks are: (i) the algorithm image registry, where the images of the \gls{ml} algorithms are stored; (ii) the secure container executor and 
computation task 
runtime components that are the libraries (Python scripts) supporting the creation of computation tasks from the \gls{ml} algorithm image; and (iii) the 
intermediate 
and completed models that are the outcome artifacts produced by the learning process which are stored in the ledger.

The following flow describes the interactions among the different building blocks and relate to the cycle of creating a \gls{ml} algorithm image, executing 
this 
image securely on learning nodes, and sharing the insights:

The image for the \gls{ml} algorithm, intended to be run as a particular instance of a federated learning process, is stored in the algorithm image registry 
(which 
can be either a shared or a private image repository). The metadata describing the \gls{ml} algorithm image, specifically an identifier of the respective 
artifact in 
the external repository and a cryptographic fingerprint (i.e., a hash value) that can be used to verify the integrity of the artifact are stored in the 
\gls{bc} ledger using the image chaincode.

During the instantiation of the execution phase, learning process metadata is created using the learning process orchestration chaincode. This metadata 
includes the unique ID for the learning process, the algorithm image this process is intended to execute, the consortium of organizational nodes participating 
in the federated learning process, indicators of the current state of the learning process (e.g., current iteration in case of iterative learning process), and 
current execution status. After the learning process metadata record is created on the chain, the \gls{ml} algorithm image is pulled from the algorithm image 
registry 
and instantiated as computation task runtime on each learning node by the secure container executor. The task is instantiated with the \gls{ml} algorithm 
runtime, 
the parameters for the run, and the initial state model. 

During the algorithm execution phase, the learning node reads the relevant insights from previous rounds by the learning nodes, runs the algorithm image on the 
relevant inputs (the insights from previous rounds, the input parameters, and the organizational datasets), and publishes the resulting insights or completed 
model to the ledger using the execution record and the model chaincodes. Once the learning process is completed, the status of the learning task on chain is 
updated to completed. 

As shown in Figure~\ref{fig:bc_figure_2}, our \gls{bc} solution comprises four chaincodes: the image and execution record chaincodes which form the secure 
execution module (Figure~\ref{fig:bc_figure_1}), and the learning process orchestration and model chaincodes which form the federated learning artifacts store 
module (Figure~\ref{fig:bc_figure_1}). The image chaincode 
provides the functionality for storing and retrieving the \gls{ml} algorithm image metadata. This chaincode also provides queries helpful in determining the 
provenance of the image, e.g., who is the creating organization or when the image was created. The learning process orchestration chaincode records the 
information about the federated learning task, including the definition of the algorithm image the learning task is about to execute; the consortium of 
organizations participating in the learning process and nodes which will run the computation tasks; and the current status of the learning process (current 
iteration, completion status). The execution record chaincode stores the execution task metadata in the ledger. Once the outcome of a single-step computation 
task, or of the particular learning round task (for iterative learning algorithms) is completed at a node, it publishes the insights to the chain (the 
estimators in our case of a fraud detection algorithm) updating the execution record. It also updates the model chaincode in case the learning process is 
completed. The model chaincode is responsible for publishing the completed model to the ledger once the learning task finishes. The complete model record 
contains a list of organizations allowed to access these models which initially equals the consortium members. 

As stressed before, one of the built-in core properties of \gls{bc} platforms is an immutable chain of blocks of transactions, establishing verifiable and 
transparent history of updates for each artifact stored in the chain. This is of fundamental importance when striving for trust and transparency of the 
execution of federated learning scenarios. Proven, verifiable, and immutable audit trail of execution tasks producing federated learning models can help 
establish without doubt, for example, that the models are derived from the desired \gls{ml} algorithm, the specific version of the algorithm, and the executing 
organizations. To this end, the artifacts usage audit logical module in Figure~\ref{fig:bc_figure_1} supports provenance for algorithm images, computation tasks 
executed, and for 
model metadata. 


\section{Experiments}\label{sec:experiments}
The implementation of the algorithm and the code to replicate the experiments presented in this section is available on 
GitHub\footnote{\url{https://github.com/GabrieleSantin/federated_fraud_detection}}.

We test the algorithm on a benchmark dataset for fraud detection\footnote{\url{https://www.kaggle.com/mlg-ulb/creditcardfraud}}. This dataset collects 
electronic credit card transactions that have been executed in some European banks during September 2013. Each transaction is represented by $28$ numerical 
features which are obtained after applying a Principal Component Analysis (PCA) on the original features, in order to hide any sensitive information, and it is 
labeled either as normal or as a fraud. The dataset contains $284807$ transactions, of which $492$ (the $0.17\%$) are frauds, making the dataset highly 
unbalanced. 

We simulate a scenario with $N:=20$ agents, each holding its own private data. To create a suitable setup, we split the given dataset into $N$ disjoint subsets 
by random sampling. To make the problem more challenging and interesting for the testing of a federated scenario, we perform an unbalanced sampling: instead 
of splitting the positive and negative examples into $N$ groups of $284807/N$ agents, we allow for each group to contain up to $70\%$ more or less 
elements than the average. Additionally, each of the resulting datasets is split into a train dataset ($90\%$ of the samples) and test dataset.
The actual number of samples for each node and the corresponding statistics are reported in Table~\ref{tab:dataset_stats}. To simplify the measurement of the 
performances of the algorithm, we artificially create a unique and centralized test set obtained by joining the $N$ test sets of the single nodes, so that all 
the test metrics are computed on the same test set. This breaks the absence of centralized orchestration in the design of the algorithm, but it is only a 
convenience choice made for the purpose of exposition.

\begin{figure}[ht]
\centering
\subfloat[{}\label{fig:topology_disconnected}]{
\includegraphics[width=0.25\textwidth]{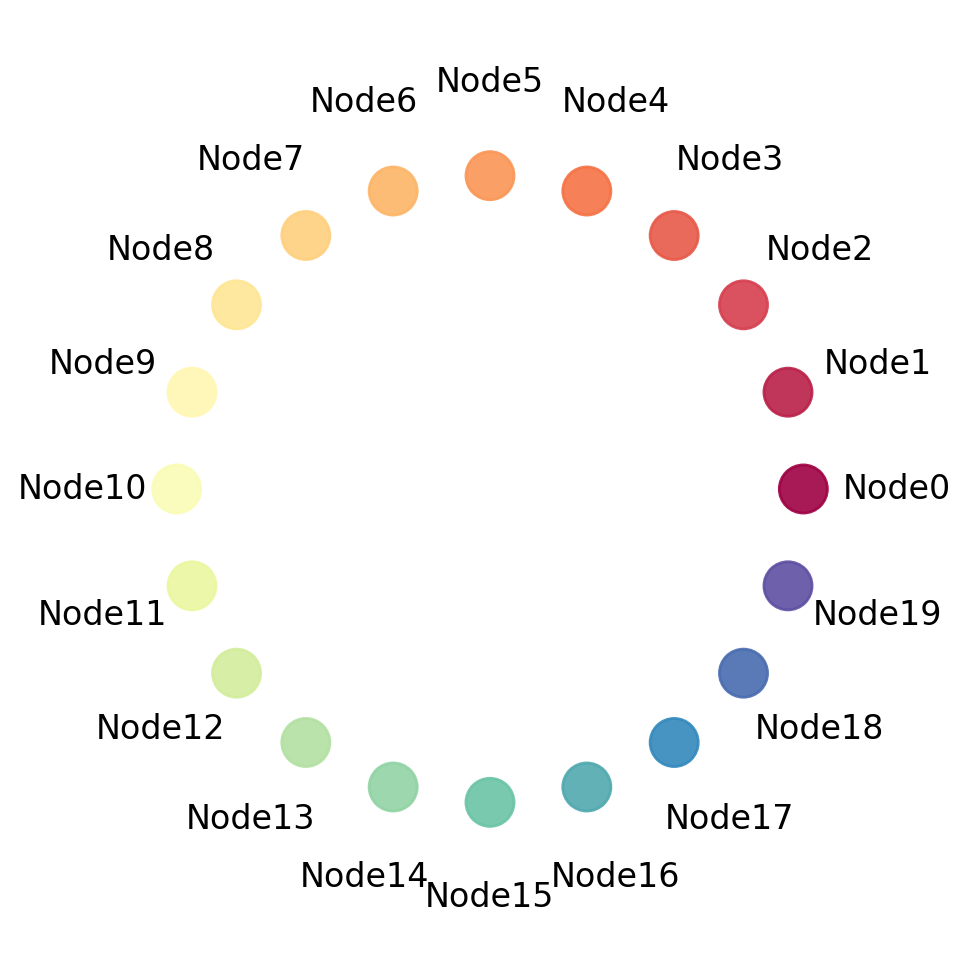}
}
\subfloat[{}\label{fig:topology_loop}]{
\includegraphics[width=0.25\textwidth]{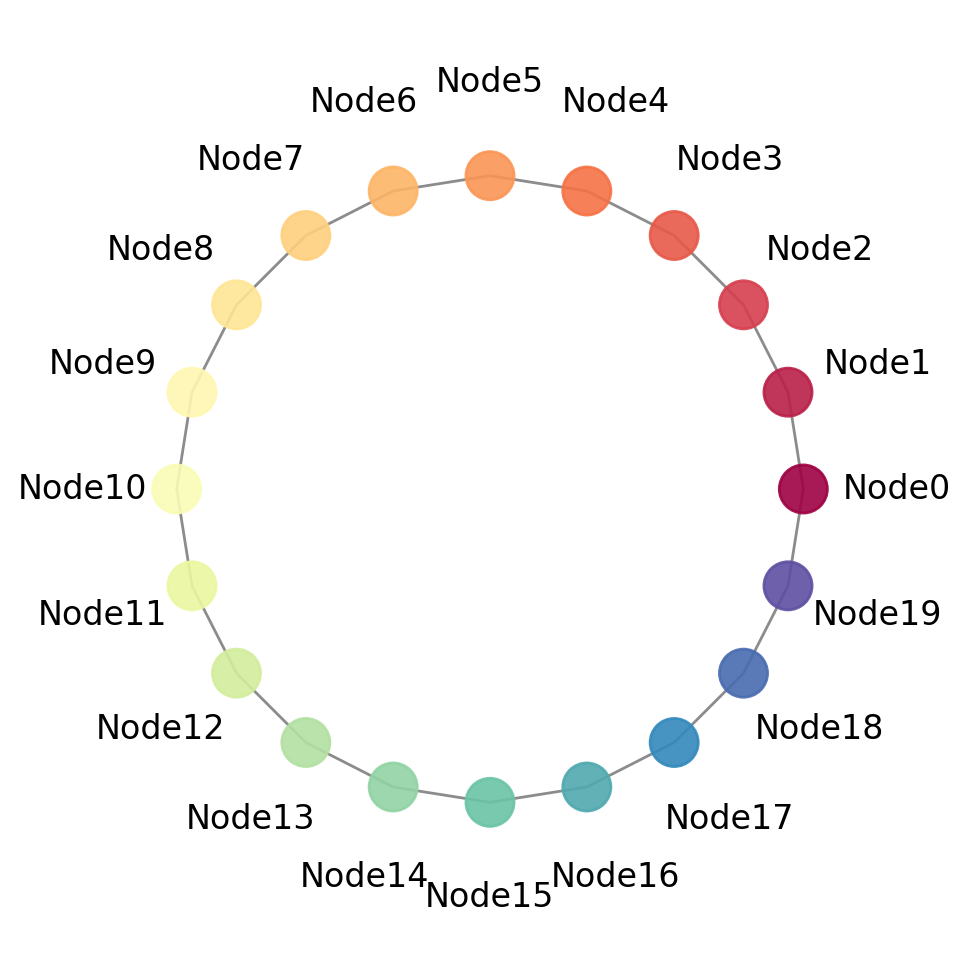}
}\\
\subfloat[{}\label{fig:topology_complete}]{
\includegraphics[width=0.25\textwidth]{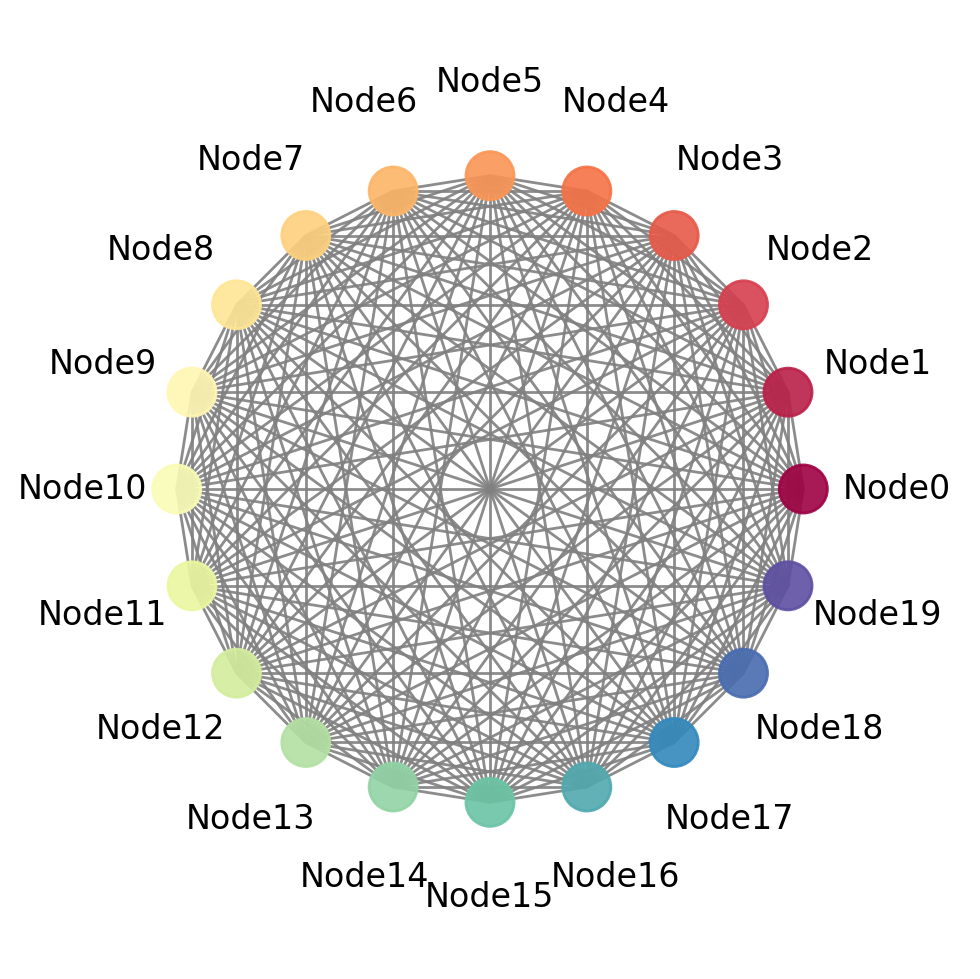}
}
\caption{Connection configurations tested in the experiments: disconnected (Figure~\ref{fig:topology_disconnected}), pairwise connected 
(Figure~\ref{fig:topology_loop}), fully connected (Figure~\ref{fig:topology_complete}).}\label{fig:topology}
\end{figure}

\begin{table}[ht]
\centering
\begin{tabular}{|ccccc|}
\hline
ID & Type & Samples & Frauds & Fraud ratio \\
\hline
Node00& Train &  14733  &     16 &      0.0011 \\
& Test  &  28489  &     57 &      0.0020 \\
\hline
Node01& Train &   9570  &     37 &      0.0039 \\
& Test  &  28489  &     57 &      0.0020 \\
\hline
Node02& Train &  12992  &      0 &      0.0000 \\
& Test  &  28489  &     57 &      0.0020 \\
\hline
Node03& Train &  15544  &     49 &      0.0032 \\
& Test  &  28489  &     57 &      0.0020 \\
\hline
Node04& Train &  13064  &     21 &      0.0016 \\
& Test  &  28489  &     57 &      0.0020 \\
\hline
Node05
& Train &  16036  &     13 &      0.0008 \\
& Test  &  28489  &     57 &      0.0020 \\
\hline
Node06
& Train &  11149  &      8 &      0.0007 \\
& Test  &  28489  &     57 &      0.0020 \\
\hline
Node07
& Train &  17571  &     32 &      0.0018 \\
& Test  &  28489  &     57 &      0.0020 \\
\hline
Node08
& Train &   3297  &     11 &      0.0033 \\
& Test  &  28489  &     57 &      0.0020 \\
\hline
Node09
& Train &  10820  &     34 &      0.0031 \\
& Test  &  28489  &     57 &      0.0020 \\
\hline
Node10
& Train &  18365  &     17 &      0.0009 \\
& Test  &  28489  &     57 &      0.0020 \\
\hline
Node11
& Train &   7875  &     33 &      0.0042 \\
& Test  &  28489  &     57 &      0.0020 \\
\hline
Node12
& Train &   8298  &      9 &      0.0011 \\
& Test  &  28489  &     57 &      0.0020 \\
\hline
Node13
& Train &  28249  &     28 &      0.0010 \\
& Test  &  28489  &     57 &      0.0020 \\
\hline
Node14
& Train &  11155  &     32 &      0.0029 \\
& Test  &  28489  &     57 &      0.0020 \\
\hline
Node15
& Train &   3363  &     12 &      0.0036 \\
& Test  &  28489  &     57 &      0.0020 \\
\hline
Node16
& Train &   7894  &     25 &      0.0032 \\
& Test  &  28489  &     57 &      0.0020 \\
\hline
Node17
& Train &  13798  &      9 &      0.0007 \\
& Test  &  28489  &     57 &      0.0020 \\
\hline
Node18
& Train &  14098  &     41 &      0.0029 \\
& Test  &  28489  &     57 &      0.0020 \\
\hline
Node19
& Train &  18450  &     11 &      0.0006 \\
& Test  &  28489  &     57 &      0.0020 \\
\hline
\end{tabular}
\caption{Size of the datasets for the 20-nodes simulation, and corresponding numbers and ratio of frauds.}\label{tab:dataset_stats}
\end{table}
To analyze the effect of different configurations of the federation, we analyze three different connection scenarios (see Figure~\ref{fig:topology}): (i) a fully 
disconnected setting, (ii) a pairwise connected setting (i.e., each node is connected to exactly two nodes), and (iii) a fully connected setting. The disconnected case 
serves as a baseline, since it represents the case where no federation takes place and each node can only rely on its own dataset.

For each configuration, we train the federated algorithm by letting each node execute the same sequence of operations (see Section \ref{sec:federated}). 
Namely, in the base case of the disconnected topology we run four repetition of \texttt{FIT}, i.e., each node creates its own model and refines it three times. 
In the connected cases, instead, we add a \texttt{SHARE} and \texttt{GET} operation after each fit. In this way, after each training on the local dataset each 
node shares its insights to its neighbors, and subsequently reads and incorporates the knowledge received by the neighbors themselves. 
The algorithm is run with values $n_{\new}:=10$, $n_{\share}:=10$, $n_{\max}:=50$ for the parameters defined in Section \ref{sec:federated}.

To measure the efficacy of the models we use three metrics, namely the balanced accuracy $\mathrm{BAcc}$, the precision $\mathrm{Prec}$, and the recall 
$\mathrm{Rec}$. Given the true test labels and the predicted test labels, we may count the number of false positive $FP$, true positive $TP$, false negative 
$FN$, false positive $FP$. With these numbers, the three metrics are defined as
\begin{align*}
\mathrm{Prec}:=\frac{TP}{TP+FP},\ \ 
\mathrm{Rec}:=\frac{TP}{TP+FN},\\
\mathrm{BAcc}:=\frac{1}{2}\left(\frac{TP}{TP+FN} + \frac{TN}{TN+FP}\right).
\end{align*}
It should be noted that all the metrics have value in $[0,1]$.

We use these metrics to assess the improvement of the federated models over the scenario where each node is isolated. To this end, for each 
node we compute on the test set the metrics in the two federated cases (pairwise connected and fully connected) and their difference with the corresponding 
value in the disconnected case. We report in Table~\ref{tab:error_max_min} the nodes for which these differences are maximal and minimal, and 
the corresponding values. It should be noted that for some nodes there is indeed a negative improvement, which means that the participation in the federation 
has a negative effect, but the corresponding values are of order at most $10^{-2}$. This is expected since the algorithm has a randomization component, and a 
change of this order of magnitude may be considered as a reasonable fluctuation. On the other hand, the maximal improvement is of order $10^{-1}$. In all 
connection scenarios and for all metrics, the node of maximal improvement is Node2: looking at Table~\ref{tab:dataset_stats}, it appears that this node has no 
frauds in the training set, and it is thus not capable of learning any meaningful classifier when isolated. On the other hand, participating in the federation 
it receives insights from its neighbors, and it is able to improve its model in a very significant way, up to an improvement of $0.9$ for the $\mathrm{Prec}$ 
metric.

\begin{table}
\centering
\begin{tabular}{|l|cccc|}

\hline
& \multicolumn{4}{c|}{BAcc}\\
& \multicolumn{2}{c}{$\min$} & \multicolumn{2}{c|}{$\max$}\\
\hline
Fully connected  &Node18 &-1.75e-02& Node2 &3.95e-01\\
Pairwise             & Node7 & 3.52e-05&  Node2&  3.95e-01\\
\hline
\hline
&\multicolumn{4}{c|}{Prec}\\
& \multicolumn{2}{c}{$\min$} & \multicolumn{2}{c|}{$\max$}\\
\hline
Fully connected & Node10& -5.22e-02  &  Node2 & 9.18e-01\\
Pairwise            & Node10& -7.93e-02  &  Node2 & 9.38e-01\\
\hline
\hline
&\multicolumn{4}{c|}{Rec}\\
& \multicolumn{2}{c}{$\min$} & \multicolumn{2}{c|}{$\max$}\\
\hline
Fully connected  & Node9 &-3.51e-02 &  Node2 & 7.89e-01\\
Pairwise             &Node3 & 0&   Node2 & 7.89e-01\\
\hline
\end{tabular}
\caption{Minimal and maximal improvement with respect to the disconnected case for the two federated scenarios 
(Fully connected and Pairwise), as measured by the three test metrics.}
\label{tab:error_max_min}
\end{table}

\begin{figure*}[ht!]
\centering
\subfloat[{}\label{fig:error_train}]{
\includegraphics[width=.45\textwidth]{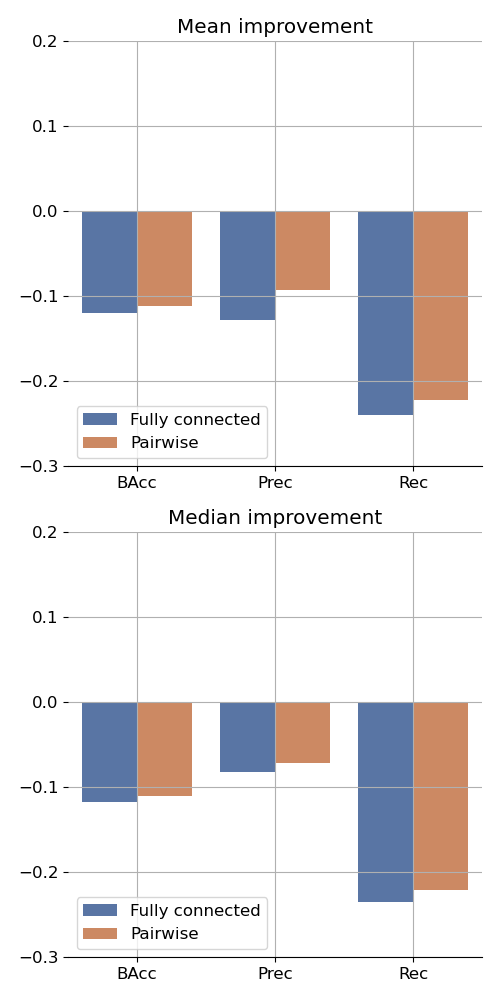}
}
\subfloat[{}\label{fig:error_test}]{
\includegraphics[width=.45\textwidth]{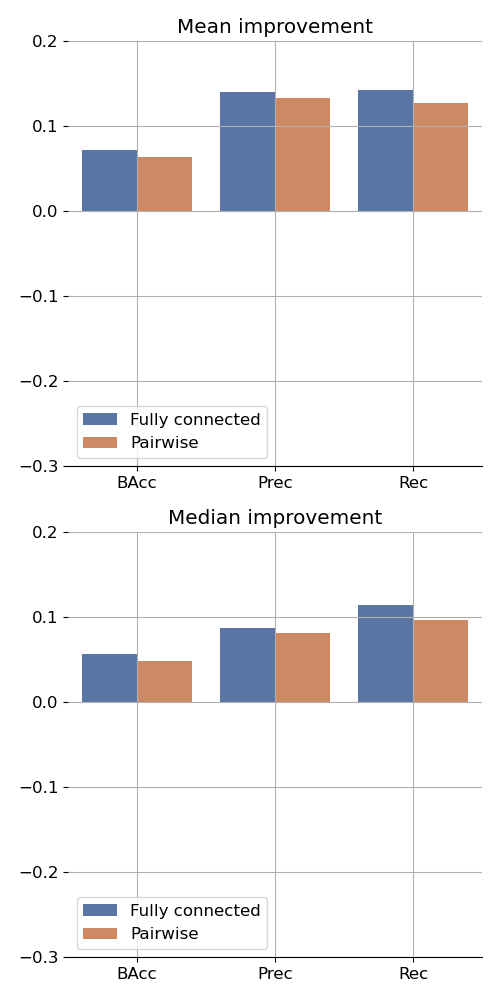}
}
\caption{Mean and median improvement in the three metrics over the disconnected case for the two federated scenarios (Fully connected and Pairwise). 
The metrics are computed over the train set (Figure~\ref{fig:error_train}) and the test set (Figure~\ref{fig:error_test}).}
\label{fig:error}
\end{figure*}

Apart from these extreme values, we compute the mean and median of these differences over the $20$ nodes. These values are reported in 
Figure~\ref{fig:error_test}, and it can be observed that overall there is a significant increase ($0.1-0.2$) both in the mean and the median, and 
for all the three metrics. This confirms that, apart from the case of single nodes, the federation is very effective to improve the 
classifiers. 

To offer an additional insight into the functioning of the sharing mechanism, we visualize in Figure~\ref{fig:error_test} the same metrics, but computed over 
the train sets of each single node. In this case, it is remarkable to observe that both the mean and median are negative, meaning that the accuracy is 
decreasing on the train set when entering the federation. Since the test metrics are instead increasing, this is a good sign that the federated algorithm is 
able to equip each node with a model that has an accuracy that goes far beyond the own dataset, and is effectively able to share insights not present in each 
single node. 

\begin{figure*}[ht!]
\centering
\includegraphics[width=\textwidth]{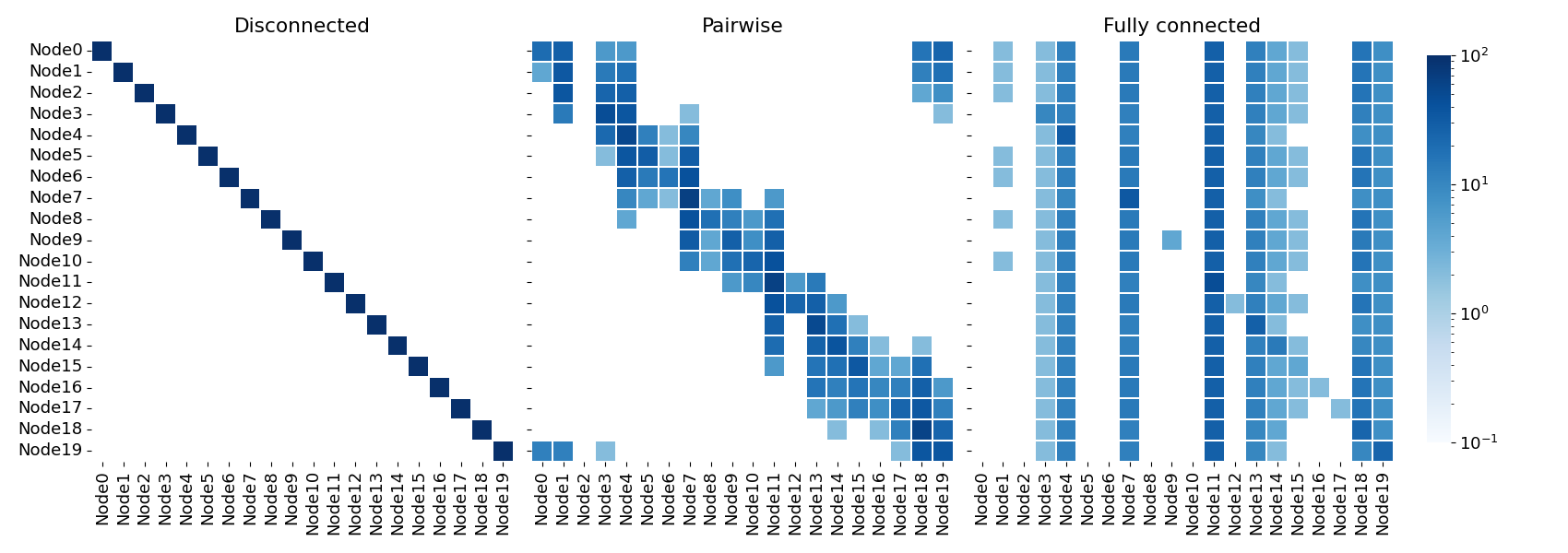}
\caption{Origin of the estimators selected by each node at the end of the iteration for the three connection settings. Each row represents a node, and the 
columns indicate the origin of its estimators. The values of each row are normalized as percentages which sum to 100\%.}
\label{fig:count}
\end{figure*}

All these results make it clear that the benefit of the federation is increased for the fully connected scenario, as one may reasonably expect. On the other 
hand, the pairwise connected setting is almost as effective. This fact is interesting in possible real applications since one may foresee that 
establishing and utilizing a connection may be expensive in different terms, and thus the nodes should be interested in establishing the minimal set of 
connections that are sufficient to obtain the desired improvement in the model. In more general terms, the effect of the topology of the connections on the 
outcome of the algorithm is an interesting aspect to explore. As a first element to explain the quite good effectiveness of the pairwise interaction, we show 
in Figure~\ref{fig:count} the distribution of the estimators over the $N=20$ nodes at the end of the iteration. Namely, since each estimator is uniquely 
identified, it is possible at each moment to check where the estimators of each node have been fitted. In the figure, we show in each row the origin of the 
estimators of each node. In the disconnected case (left panel) there is no mix, and indeed each node owns only estimators that it fitted itself. In the 
fully connected case (right panel) a quite uniform mixing can instead be observed, with the addition that some nodes (Node0, Node2, Node5, Node6, Node8, 
Node10) produce almost no estimators that are used by the other ones. The fact that the mixing is quite stable among the nodes is an indication of the 
effectiveness of the sharing and ranking mechanism. In the intermediate case of the pairwise connected nodes (central panel) the mixing reflects the connection 
pattern, since each node holds estimators from its direct neighbors. In this case it is worth remarking that the estimators are effectively transmitted beyond 
the first order neighbors of a node, and this suggests that even a not fully connected network may be effective for the federation to work.

\section{Conclusions and future work}\label{sec:conclusions}
In this paper we developed and presented a federated anomaly detection algorithm that can leverage the communication between collaborating nodes in order to 
improve the models' performances. The algorithm is designed according to a fully decentralized structure, and it allows the sharing of algorithmic insights 
without the movement of any data. Although the classifiers are defined on top of Random Forests, we discussed how the same structure can be adapted to more 
general scenarios.

Remarkably, the federated learning algorithm is developed in order to be fully integrated into a \gls{bc} solution that ensures privacy-preserving guarantees 
on the execution and on its results. This component make it possible to verify the identity of the 
participating nodes, and to audit the execution of the algorithms and the correct functioning of the federation. Also in this case, the \gls{bc} solution is 
not bounded to the specific algorithm of choice, and we discussed how more general machine learning models may be secured within the same framework.

Extensions of the basic algorithmic structure will be analyzed in future work, where more complex building blocks can be exploited in place of Random Forests 
and Decision Trees. Moreover, the effect of the connection topology on the behavior of the algorithm has been only partially explored in this work, and 
interesting options for its optimization remain open.
Ultimately, we may foresee the application of these techniques to leverage the models stored in the \gls{bc} for data sharing and trading in a data 
marketplace. Data marketplaces for \gls{ml} models are an emerging trend \cite{Zyskind2015,Nasonov2018,Ha2019,Travizano2020}, which provide the opportunity to decentralize model development and lower the entry 
barrier into \gls{ml} usage for companies which do not have  either the skills, the capacity, or the access to learning data to develop the algorithms and train 
the models. 
Chaincodes in the \gls{bc} network could control the access and permissions to the different models stored in chain applying governance rules defined by the consortium organizations.

\section*{Acknowledgments}
The work of the authors was partially supported by the H2020 INFINITECH project, grant agreement number 856632.

\bibliographystyle{abbrv}
\bibliography{biblio}

\appendix
\section{Construction of the tree kernel}\label{sec:kernel}
We consider a set $\mathcal D:=\{D_i\}_{i=1}^{n_D}$ of $n_D\in\N$ decision trees, where $D_i:=(T_i, X_i)$, $T_i$ is a tree where each non-terminal node $v$ has 
a label $s(v)\in\{1,\dots, d\}$, $d\in\N$, and for each node $x(v)\in\R$ is node feature.

We define a positive definite and symmetric kernel over $\mathcal D$ by a modification of the convolutional kernel of \cite{Haussler1999,Collins2001}. 
Namely, we first enumerate the set $t_1, \dots, t_{M}$ of all subtrees of the trees in $\mathcal D$. We remark that the trees here are labeled, meaning that 
the trees are equal only if the corresponding nodes have the same label. Given a tree $T$ and any node $v\in T$, we then define a feature map 
\begin{equation*}
h(v):=\left[I_1(v), \dots, I_M(v)\right]^T\in \{0,1\}^M,
\end{equation*}
where $I_i(v) = 1$ if and only if the subtree $t_i$ is rooted in $v$. This allow us to define the kernel $k_{\node}$ of \cite{Collins2001} between two 
nodes $v\in T$, $v\in T'$, as
\begin{equation*}
k_{\node}(v, v'):= h(v)^T h(v') = \sum_{i=1}^M h_i(v) h_i(v').
\end{equation*}
It can be proven that $C(v, v')$ can be efficiently computed in polynomial time, and it simply counts the number of common subtrees rooted at both $v$ and 
$v'$ (see \cite{Collins2001}). This kernel can be used to define a tree kernel $k$ between $T, T'$ simply by aggregation over all pairs of nodes, i.e., 
\begin{equation*}
k_{\tree}(T, T'):= \sum_{v\in T, v'\in T'} k_{\node}(v, v').
\end{equation*}

We extend this definition to a kernel $k$ on our Decision Trees $D:=(T, X), D':=(T', X')\in\mathcal D$ simply by adding a second kernel that takes into account 
the values of the node features, namely we sets
\begin{equation*}
k(T, T'):= \sum_{v\in T, v'\in T'}  k_{\feat}(x(v), x(v')) k_{\node}(v, v'),
\end{equation*}
where $k_{\feat}:\R\times\R\to\R$ is any positive definite kernel. Observe that $k$ is positive definite because it is obtained by sums and products of 
positive definite kernels \cite{Aronszajn1950}. Moreover, for simplicity we use the linear kernel $k_{\feat}(x(v), x(v')):= x(v) x(v')$, and this make it 
possible to write also $k$ as an aggregation over node kernels via 
\begin{align*}
k(T, T')
&= \sum_{v\in T, v'\in T'}k_{\feat}(x(v), x(v')) k_{\node}(v, v') \\
&= \sum_{v\in T, v'\in T'}x(v) x(v') h(v)^T h(v') \\
&= \sum_{v\in T, v'\in T'}h_x(v)^T h_x(v'),
\end{align*}
where $h_x(v):=\left[x(v) I_1(v), \dots, x(v)I_M(v)\right]^T\in \R^M$.

\end{document}